\documentclass[sigconf]{acmart}
\makeatletter
\renewcommand\@formatdoi[1]{\ignorespaces}
\makeatother

\usepackage{booktabs} 

\setcopyright{rightsretained}

\usepackage{pgfplots}
\usepackage{multirow}
\usepackage{amsmath}
\usepackage{subcaption}
\usepackage{enumitem}
\usepackage{graphicx}

\newcommand{\NAME}[0]{KnowNER}

\newcommand\eat[1]{}

\newcommand{\specialcell}[2][c]{%
	\begin{tabular}[#1]{@{}c@{}}#2\end{tabular}}

\acmDOI{}

\acmISBN{}

\begin{document}
\title{\NAME: Incremental Multilingual Knowledge in Named Entity Recognition}


\author{Dominic Seyler}
\affiliation{
	\institution{University of Illinois at Urbana-Champaign}
}
\email{dseyler2@illinois.edu}

\author{Tatiana Dembelova}
\affiliation{
	\institution{Max Planck Institute for Informatics}
}
\email{tdembelo@mpi-inf.mpg.de}

\author{Luciano Del Corro}
\affiliation{
	\institution{Max Planck Institute for Informatics}
}
\email{corrogg@mpi-inf.mpg.de}

\author{Johannes Hoffart}
\affiliation{
	\institution{Max Planck Institute for Informatics}
}
\email{jhoffart@mpi-inf.mpg.de}

\author{Gerhard Weikum}
\affiliation{
	\institution{Max Planck Institute for Informatics}
}
\email{weikum@mpi-inf.mpg.de}

\begin{abstract}
\NAME{} is a multilingual Named Entity Recognition (NER) system that leverages different
degrees of external knowledge. A novel modular framework divides
the knowledge into four categories according to the depth of knowledge they
convey.  Each category consists of a set of features automatically generated from
different information sources (such as a knowledge-base, a list of names 
or document-specific semantic annotations) and is used to train a conditional random field
(CRF). Since those information sources are usually multilingual,
KnowNER can be easily trained for a wide range of languages. 
In this paper, we show that the incorporation of deeper knowledge systematically
boosts accuracy and compare \NAME{} with state-of-the-art NER
approaches across three languages  (i.e., English, German and Spanish) performing amongst 
state-of-the art systems in all of them. 
\end{abstract}

%
%



\maketitle

\section{Introduction}

Named Entity Recognition (NER) is the task of detecting named entity mentions in text and assigning them to their corresponding
coarse-grained type (e.g., person, location, organization, miscellaneous). For instance, given the sentence ``Jimmy Page played in New York'', the goal is to recognize ``Jimmy Page'' and ``New York'' as named entities and classify them as person and location. NER is a key component in a wide range of natural language understanding tasks such as named entity disambiguation (NED), information extraction, question answering, machine translation, knowledge graph construction, etc.

Here we present \NAME{}, a multilingual NER system which incorporates different degrees of external knowledge through language agnostic features, designed to exploit existing multilingual knowledge resources.

In contrast to previous approaches, \NAME{} is implemented as a modular framework, drawing on different sources of external knowledge. 
We divide the information sources into four different categories according to the depth of knowledge they convey.
Each one carries more information than the previous.
This additional knowledge boosts accuracy but also increases the processing overhead,
establishing a clear accuracy-speed trade-off that can be exploited according to processing requirements and
the availability of computational and knowledge resources.

This work has three main goals: (i) present a high performance knowledge intensive NER framework that can be used for a wide range of languages, (ii) understand to which extent external knowledge improves NER performance, and (iii) present a novel set of knowledge intensive features that can be used in a multilingual setting.

\NAME{} implements a linear chain CRF, which was proven to work well for the NER task~\cite{FinkelGM05}. We divide the features according to the knowledge categories defined below:
 
\noindent\textbf{Agnostic.} These features correspond to the standard lexico-syntactic features extensively used in literature~\cite{FinkelGM05}. They are usually called local features since they are directly extracted from text and do not use any external knowledge. For instance, part-of-speech (POS) tags are good indicators of named entities (e.g., In ``Jimmy Page plays guitar'', ``Jimmy'' and ``Page'' are proper nouns).

\noindent\textbf{Name-based.} This set is extracted from a list with millions of named entity names. They unveil common patterns or attributes that indicate the presence of named entities. For example, the word``Jimmy'' is usually associated with named entities.

\noindent\textbf{KB-based.} This group is generated from a knowledge base (KB) or an entity annotated corpus. The aim is to go beyond the surface forms exposing particular semantics of the named entities (e.g., their types). Following Ratinov and Roth.~\cite{RatinovR09}, we use gazetteers that associate named entity names with types. We generate them in an automatic way from YAGO~\cite{SuchanekKW07}, a multilingual KB. We also use it to extract richer information like the probability of a single token having a given type and appear in a specific position (e.g., the probability of ``Jimmy'' being a person and appearing at the beginning of the name). Additionally, we exploit an annotated corpus to estimate the likelihood of a token referring to a named entity (e.g., The number of times ``Page'' is linked in Wikipedia articles).
 
\noindent\textbf{Entity-based.} These features exploit information from a particular document. The idea is that if some entities in the document are identified in advance, it is easier to spot more difficult cases later. For instance, if we know that the \textit{European Union} is mentioned in a text, we can assume that the token ``EU'' will most probably  refer to it. Previous work~\cite{RadfordCH15} builds on this idea, using disambiguated entities from ground truth data to extract document specific features. We follow this approach but, in addition, we evaluate our system in a real world scenario using AIDA~\cite{Hoffart:2011a}, a state-of-the-art entity-linking system.

When \NAME{} includes all knowledge categories, it performs among the best NER systems across all the evaluated languages (i.e., English, German and Spanish) on four standard datasets. In the experimental section, we also present an extensive study showing that the degree of knowledge correlates positively with task accuracy and negatively with processing time. We also show that external knowledge is particularly important for types like organizations, persons, and locations, reaching in the last two cases human-level accuracy (more than 95 F1 points) for the English language. Apart from the traditional NER metrics (class label plus text span) we additionally report the span recognition accuracy (named entity without the type tag), essential for certain tasks (e.g., NED).

Now we summarize our central contributions:

\begin{itemize}[noitemsep]
\item A high performance multilingual NER system based on a modular framework for incorporating different types of external knowledge.	
\item A comprehensive study to verify the impact of external knowledge into NER, including ablation and timing experiments.
\item A multilingual set of knowledge intensive automatically generated features derived from large list of names, or a multilingual KB.
\item Real world scenario experiments to test the specific effects of NED into NER. 

\end{itemize}

\section{The Named Entity Recognition Task}
\subsection{Task Definition} 
\label{sub:task_definition}

The goal of NER is to find named
entity mentions in text and map them to pre-defined types (e.g., person,
location, etc.). For instance, in the sentence ``Jimmy Page plays guitar.''
the goal is to recognize that the text span ``Jimmy Page'' refers to a named entity that can be categorized as person.

The task implies two challenges: (i) Find the text span of a named entity name 
and (ii) Annotate each named entity with a type.
The first challenge requires identifying tokens that refer to named entities. A named entity may be composed by more than one token (``United States''), and a named entity may be embedded in another named entity (``Supreme Court of the United States''). 
The second challenge requires deeper semantic understanding (e.g. understand
that ``Jimmy Page'' is not only a named entity but specifically a person). 

Although NER commonly refers to both tasks, some applications may rely only on the first one (e.g. NED). 
In Sec.~\ref{sec:evaluation} we present 
results for the named entity mention span detection separately. 

\subsection{A linear chain CRF model}

Previous work~\cite{FinkelGM05,KazamaT07a,RatinovR09,PassosKM14,RadfordCH15,LuoHLN15} proved the effectiveness of CRFs~\cite{Lafferty01} 
for the NER task. We implemented \NAME{} as a linear chain CRF similar to \cite{FinkelGM05}. 
The underlying idea is to cast NER as a sequence model with a bidirectional flow. The CRF represents the probability of a hidden state sequence (i.e., token labels) given a set of observations.  In a linear chain CRF, the probability of a token being a named entity depends on a set of observations including the label of its adjacent neighbors. For a more in-depth description of the model refer to Finkel et al., 2005.

\begin{table*}[t!]
    \begin{center}
        \scalebox{0.9}{
        \begin{tabular}{|c||l|l|l|}
            \hline
            \bf Cat. & \bf Feature & \bf Description & \bf Example\\
            \hline \hline
            \multirow{5}{*}{A}
            & Word & specific words tend to indicate the presence of a NE & John says that \dots $\rightarrow$   \\
            &      &                                                      & [SOMEONE] says that \dots \\
			\cline{2-4}
            & Word shapes & NE have specific shapes & John, Paul $\rightarrow$ Xxxx \\
            \cline{2-4}
            & POS tags & NE tend to have specific POS tags & John, Paul $\rightarrow$ NNP \\
            \cline{2-4}
            & Prefixes/Suffixes & NE tend to share prefixes and suffixes &  Frei\underline{burg}; Mar\underline{burg}\\
            \cline{2-4}
            & Presence Window & NEs usually don't appear twice in a small window & To be or not to be \\
			&                 &                                                  & Obama was born in Hawaii \\     
			
            \cline{2-4}
            & Sentence Begin & NE at the beginning of sentences difficult to spot & John says \dots; Computers are \dots \\
            \hline
            \multirow{2}{*}{Name}
            & \textbf{Mention tokens} & Some tokens are strongly associated to NEs & county,john,school,station,\dots   \\
            \cline{2-4}
            & \textbf{POS-tag sequence} & Multi-word NEs tend to share POS patterns & Organization of American States \\
            &  &  & Union for Ethical Biotrade  \\
			&  &  & $\rightarrow$ NNP IN NNP NNP \\
            \hline
            \multirow{3}{*}{KB}
            & Type gazetteers & Some names are strongly associated to types & Barack Obama $\rightarrow$ person\\
			&                 &                                             & Florida $\rightarrow$ location \\  
            \cline{2-4}
            & \textbf{Wiki. link prob.} & Certain tokens are usually associated to NEs & Obama is usually linked \\
			&                           &                                              & to \textit{Barack Obama} in Wikipedia \\
            \cline{2-4}
            & \textbf{Type prob.} & Certain tokens are associated to types with high probability & Barack $\rightarrow$ person; \\
            \hline
            Entity &  Doc. gazetteers & Presence of specific NEs may indicate other NE names & European Union $\rightarrow$ EU \\ \hline
        \end{tabular}
        }
        \caption{\label{table:feature-classes} Features by category (novel features are highlighted)}
    \end{center}
    \vspace*{-0.8cm}
\end{table*}

\section{Knowledge Augmented NER}
\label{sec:knowledge-aug-ner}

Here, we describe the knowledge categories, which function as modules in our system. We define four: \textit{agnostic (A)}, \textit{name-based},
\textit{KB-based} and \textit{entity-based}, each containing an increasing amount of external knowledge. A category consists of the set of features, sumarized in Tab.~\ref{table:feature-classes}.

\subsection{Knowledge Agnostic}
\label{ssec:agnostic-features}

This category contains the so-called ``local'' features. Their distinctive characteristic is that they can be extracted directly from text without
any external knowledge.
These features are mostly of a lexical, syntactic or linguistic nature and have been well-studied in literature. We implement
most of the features described in Finkel et al.~\cite{FinkelGM05} and Zhang and Johnson.~\cite{Zhang03}, namely: 

(1) The current word and words in a window of size 2 ;
(2) Word shapes of the current word and words in a window of size 2;
(3) POS tags in a window of size 2;
(4) Prefixes (length three and four) and Suffixes (length one to four); 
(5) Presence of the current word in a window of size 4;
(6) Beginning of sentence.

\subsection{Name-Based Knowledge}
\label{ssec:name-based-features}

In this category, the knowledge is extracted from a list of named entity names. This list does not carry any additional information apart from the names themselves. The intuition is that names tend to follow patterns and even the set of possible names is limited. To the best of our knowledge, these features have not been previously used.
We extracted a list of all names from YAGO~\cite{SuchanekKW07} (30.85M for the languages we trained on) and created the following features:

\textit{Frequent mention tokens.} Reflects the frequency of a given token in a list of entity names.
We tokenized the list to compute the frequencies. The feature assigns a weight to each token in the text corresponding to their normalized frequency. The intuition is that some words like ``John'' or ``Organization'' may be indicative of a named
entity and thus carry a high weight. For instance, the top-5 tokens we found in English were ``county'', ``john'', ``school'', ``station'' and ``district''. 
All tokens without occurences are assigned 0 weight.

\textit{Frequent POS Tag Sequences.} This feature intends to identify POS sequences common to named entities. For example,
person names tend to be described as a series of proper nouns, while organizations may have richer patterns. For instance, both
``Organization of American States'' and ``Union for Ethical Biotrade'' share the pattern NNP-IN-NNP-NNP, where NNP is a proper noun
and IN a preposition.
To generate these patterns, we construct a simple artificial sentence for each name in our list and run a POS-tagger. We then compute and rank the entity POS tag sequences and keep the top 100. 
The feature is implemented by finding the
longest matching POS sequences in the input text and marking whether the current token belongs to a frequent sequence or not. 
We search the sequences from left to right and, in case of overlap, annotate only the leftmost sequence. This might need to be done differently
for languages that read right to left.

\subsection{Knowledge-Base-Based Knowledge} 
\label{ssec:knowledge-base-features}

This category groups features that are extracted from a KB or an entity annotated corpus. They encode knowledge about named entities themselves or their usages.
Conceptually, we aim to incorporate the likelihood of a particular token being linked to an entity of a specific type. We implemented three features:

\textit{Type-infused Gazetteer Match.} It finds the longest occurring token sequence in a type specific gazetteer. It
adds a binary indicator to each token, depending on whether the token is part of a sequence. 
We use 30 dictionaries distributed by Ratinov and Roth, 2009 containing type-name information for English. For instance, ``New York'' is a place and ``McDonald's'' a corporation.
These dictionaries have been successfully used in the past~\cite{PassosKM14,RadfordCH15,LuoHLN15}. For the rest of the languages we generated the dictionaries automatically by mapping each dictionary to a set of YAGO types and extracting the corresponding names. For the dictionary containing corporations, for example, we incorporated all the names in the specific language corresponding to types \textit{company} and \textit{enterprise}. 

\textit{Wikipedia Link Probability.} This feature measures the likelihood of a
token being linked to a named entity Wikipedia page. The intuition is 
that tokens linked to named entity pages tend to
be indicative of named entities. For instance, the token ``Obama'' is usually linked while the term ``box'' is not. The list of pages referring to named entities is extracted from YAGO.  Given a token in the text, it is assigned the probability of being linked according to  Eq.~\ref{eq:wiki-links}, where $link_{d}(t)$
equals 1, if token $t$ in document $d$ is linked to another Wikipedia
document. $present_{d}$ equals 1 if $t$ occurs in $d$.

\begin{equation}
\label{eq:wiki-links}
    P_{Wiki}(t) = \dfrac{\sum_{d \in D}link_{d}(t)}{\sum_{d \in D}present_d(t)}
\end{equation}

Since usually in Wikipedia only the first occurrence of a named entity is linked, we count a word on a page as linked if it links to a named entity page at least once. 

\textit{Type Probability.} Intended to discriminate between types, it encodes the likelihood of a token belonging to a given type.
The idea is to capture the fact that, for instance, the token ``Obama'' is more likely a person than a location.
Since YAGO contains types and names for each entity,
we can calculate the conditional probability. 

Given a set of entities $E$ with mentions $M_e$
and tokens $T_{em}$ we calculate the probability
of a class $c \in C$ given a token $t$ as

\begin{equation}
\label{eq:class-type-prob}
    P(c|t)=\dfrac{\sum_{e}^{E}\sum_{m_{e}}^{M_{e}}\sum_{t_{em}}^{T_{em}} c(e)}                    
    {\sum_{e}^{E}\sum_{m_{e}}^{M_{e}}\sum_{t_{em}}^{T_{em}}\sum_{c_{i}}^{C} c_{i}(e)}
\end{equation}
where $c(e) = 1$ if entity $e$ belongs to class $c$ and $c(e) = 0$ otherwise. For each token in the text, we create one feature per type with the respective probability as its value.

\textit{Token Type Position.} Attempts to reflect that tokens
may appear in different positions according to the entity type. For instance,
``Supreme Court of the United States'', is an organization and 
``United'' occurs at the end. In ``United States'', a location,
occurs at the beginning. This helps
with named entities inside other named entities.

This idea is implemented using the BILOU (\textbf{B}egin, \textbf{I}nside, \textbf{L}ast,
\textbf{O}utside, \textbf{U}nit) encoding~\cite{RatinovR09}, which tags each token with
respect to the position in which it occurs (e.g., ``\textbf{O}-The \textbf{B}-Supreme \textbf{I}-Court \textbf{I}-of \textbf{I}-the \textbf{I}-United \textbf{L}-States''). The number of features depends on the number of types in the dataset (4 BILU positions times $n$ classes + O position). For each token, each feature receives the probability of a class given the token and position.
The class probabilities are calculated as in Equation~\ref{eq:class-type-prob}, incorporating also the token position. 
This strategy gives us the possibility to combine the class type probabilities with the token positions.

To the best of our knowledge, the last three features (\textit{Token Type Position}, \textit{Type Probability} and \textit{Wikipedia Link Probability}) have not been used in previous work.

\subsection{Entity-Based Knowledge}
\label{ssec:entity-features}

This category encodes document specific knowledge about the entities found in text. 
The idea is to exploit the inherent association between NER and NED. 
Previous work showed that the flow of information between the two tasks
generates significant improvements in NER performance~\cite{RadfordCH15,LuoHLN15}.

Comparatively, this module requires more (computational and knowledge) 
resources than the previous ones. 
It requires a first run of NED to generate document
specific features, based on the disambiguated named entities. The generated features are used in a second run of NER. 

Following Radford et al.~\cite{RadfordCH15}, after the first run of NED, we create a set of document-specific gazetteers derived from 
the named entities found. The idea is that this information will help
in the second round to find new named entities missed in the first one.
Take the sentence ``Three-quarters of citizens of the European Union working in the
United Kingdom would not meet current visa requirements for non-EU overseas
workers if the uk left the bloc''. We can imagine that in the first round of NED
\textit{European Union} and \textit{United Kingdom} can be easily identified. However, ``EU'' or the wrongly
capitalized ``uk'' might be missed. After the disambiguation, we know that both disambiguated entities are
organizations and have the aliases \textit{EU} and \textit{UK}
respectively. The idea is that if we introduce this information in a second NER run, they are easier to spot.

For each document we gather all entities that were disambiguated in the first NED run. Then we extract all surface forms of the identified entities from YAGO. The surface forms are tokenized and assigned the type of the corresponding entity plus its BILOU position. For example, the surface form ``Barack Obama'' will result in the two tokens ``Barack'' and ``Obama'', which will be assigned to ``B-Person'' and ``L-Person'' respectively. In \NAME{} this feature is incorporated as 17 binary features (BILU tags multiplied by 4 coarse grained types + O tag), which fire when a token is encountered that is part of a list that contains the mappings from tokens to type--BILOU pairs.

\section{Evaluation}
\label{sec:evaluation}

In this section, we analyze the effect of external knowledge (Sec.~\ref{sec:incremental-knowledge}) and compare \NAME{} with state-of-the-art approaches (Sec.~\ref{sec:experimental-setup}) for three languages: English, German and Spanish.

\subsection{Experimental Setup}
\label{sec:experimental-setup}

\textbf{\NAME.} The CRF was trained using CRF-suite~\cite{CRFsuite} with
the Limited-memory Broyden-Fletcher-Goldfarb-Shanno algorithm and L1 regularization (coeff. = 1), which performed best on the 
English CoNLL2003 dev. set. We provide two settings for the system: (i) \NAME{}$_{gold}$, which as Radford et al.~\cite{RadfordCH15} uses the gold standard 
named entity annotations for the Entity-based features and was used to analyze the impact of knowledge into the NER task,
and (ii) \NAME{}$_{aida}$ which runs AIDA to produce the Entity-based features and was used across all languages
to compare with other available NER systems.

\textbf{Datasets.} We evaluated \NAME{} on four well established datasets that provide annotated named entity mentions and types. 

\textit{CoNLL2003e.} By Sang and Meulder~\cite{SangM03}, it is a collection of English
Reuter's newswires with named entity mentions annotated with types (i.e., persons, locations, organizations and miscellaneous). CoNLL2003e-dev does not include the developing set in training while CoNLL2003e-test does.

\textit{MUC-7.} A set of New York Times articles (in English)~\cite{Chinchor1997muc} designed for NED and NER.
It annotates named entities and their types (i.e., organizations, persons, locations), dates, times, and quantities (monetary values, percentages). We only focused on the named entity types.  MUC-7-dev does not include the developing set in training while MUC-7-test does.

\textit{CoNLL2003g.} A German dataset also by Sang and Meulder~\cite{SangM03}, similar to CoNLL2003e the named entities are classified according to four types (i.e., persons, locations, organizations and miscellaneous). It consists of a collection of news articles from the Frankfurter Rundschau.

\textit{CoNLL2002.} By Tjong Kim Sang~\cite{TjongKimSang:02}, it is a collection of news wire articles in Spanish made available by the Spanish EFE News Agency. The named entities are classified into persons, organizations, locations, times and quantities. We only focus on the first three.

\textbf{Metrics.}
We report $F_{1}$-score for all systems and two evaluation methodologies
for our system and the other methods, when available.

\textit{Mention-based.} It considers a named entity prediction as correct, if and only if the
mention boundaries and predicted types are exact matches with the gold standard.

\textit{Span-based.} Measures the correctness of mention boundaries ignoring type labels.
This measure is important for applications which do not necessarily require type annotations such as NED.

\textbf{Knowledge depth.} 
To demonstrate the impact of increasing knowledge on NER performance we tested four variations of \NAME{}, equivalent to the categories introduced in Sec.~\ref{sec:knowledge-aug-ner}. Each variation contains the features corresponding to a category name plus all those from the lighter categories.

\textit{Agnostic:} \NAME$_{A}$ uses only local lexico-syntactic features without any external knowledge resources (Sec.~\ref{ssec:agnostic-features}).
 
 \textit{Name-based:} \NAME$_{Name}$ uses features based on a list of names (Sec.~\ref{ssec:name-based-features}) plus those in \NAME$ _{A} $.
  
  \textit{KB-based:} \NAME$_{KB}$ utilizes features derived from a knowledge-base (Sec.~\ref{ssec:knowledge-base-features}) plus \NAME$ _{Name} $ features.
  
  \textit{Entity-based:} \NAME$_{Entity}$ requires the execution of NED to generate document based features (Sec.~\ref{ssec:entity-features}) in addition to all the features in \NAME$ _{KB}$.

 \begin{figure*}[!hbt]
     \centering
     \begin{subfigure}{0.48\textwidth}
        \includegraphics[width=1\textwidth,keepaspectratio]{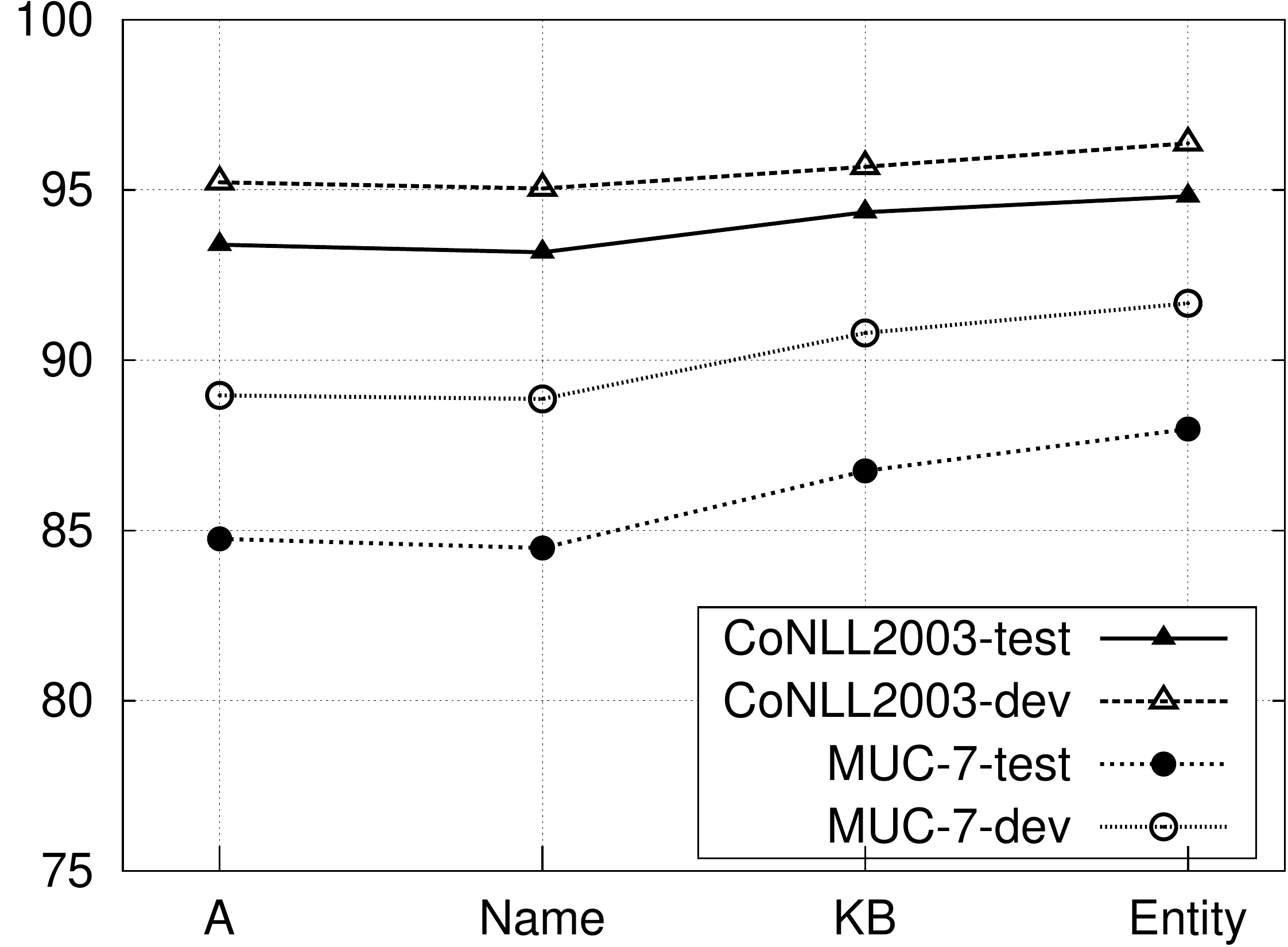}
        \caption{Span-based $F_{1}$ score}
        \label{fig:feature-categories-performance-span}
     \end{subfigure}
     \quad
     \begin{subfigure}{0.48\textwidth}
	 	\includegraphics[width=1\textwidth,keepaspectratio]{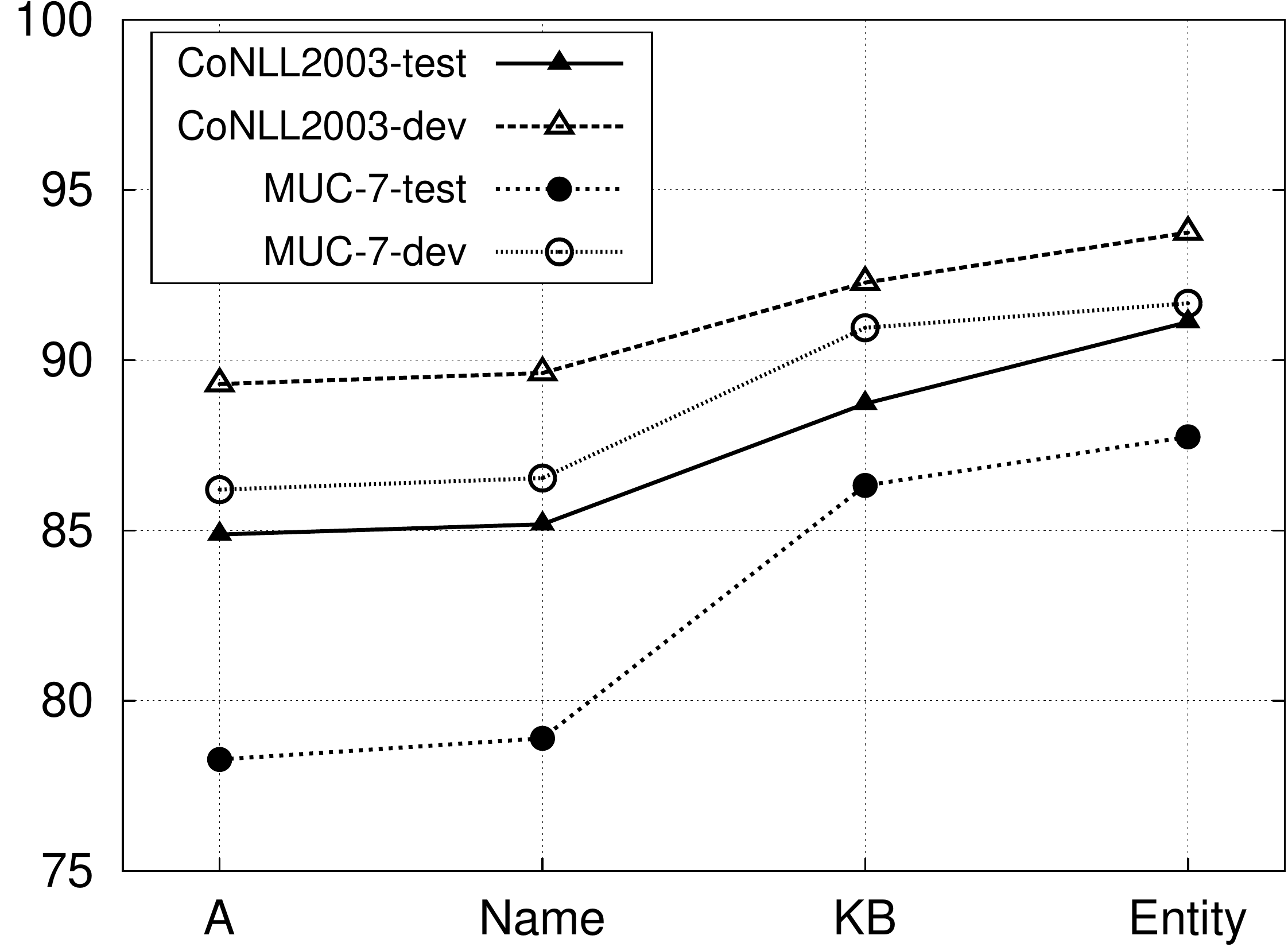}
	 	\caption{Mention-based $F_{1}$ score}
	 	\label{fig:feature-categories-performance}
     \end{subfigure}
	 \caption{\NAME{}$_{gold}$: Mention and span-based results for CoNLL2003e and MUC-7}
 \end{figure*}

\subsection{Incremental Knowledge}
\label{sec:incremental-knowledge}

Here we analyze the impact of external knowledge on the system. Our detailed analysis is specific for the 
English language on the CoNLL2003e and MUC-7 datasets. Results show a clear improvement when deeper knowledge
is used across datasets and entity types.

Fig.~\ref{fig:feature-categories-performance-span} shows the effect of the span detection in each category. Although it drops slightly for the name-based category, it quickly recovers as deeper knowledge is added. The effect is similar for both datasets.

Regarding the mention-based metric, Fig.~\ref{fig:feature-categories-performance} shows the effect of different knowledge categories
for the Mention-based metric.
In all cases adding knowledge generates a boost in performance.
The effect is particularly strong for MUC-7-test which registered an overall increment of almost 10 F1 points.
In both cases, the biggest boost is registered when the KB-based features are added.

\begin{figure}[!htb]
	 \centering
     \includegraphics[width=0.48\textwidth,keepaspectratio]{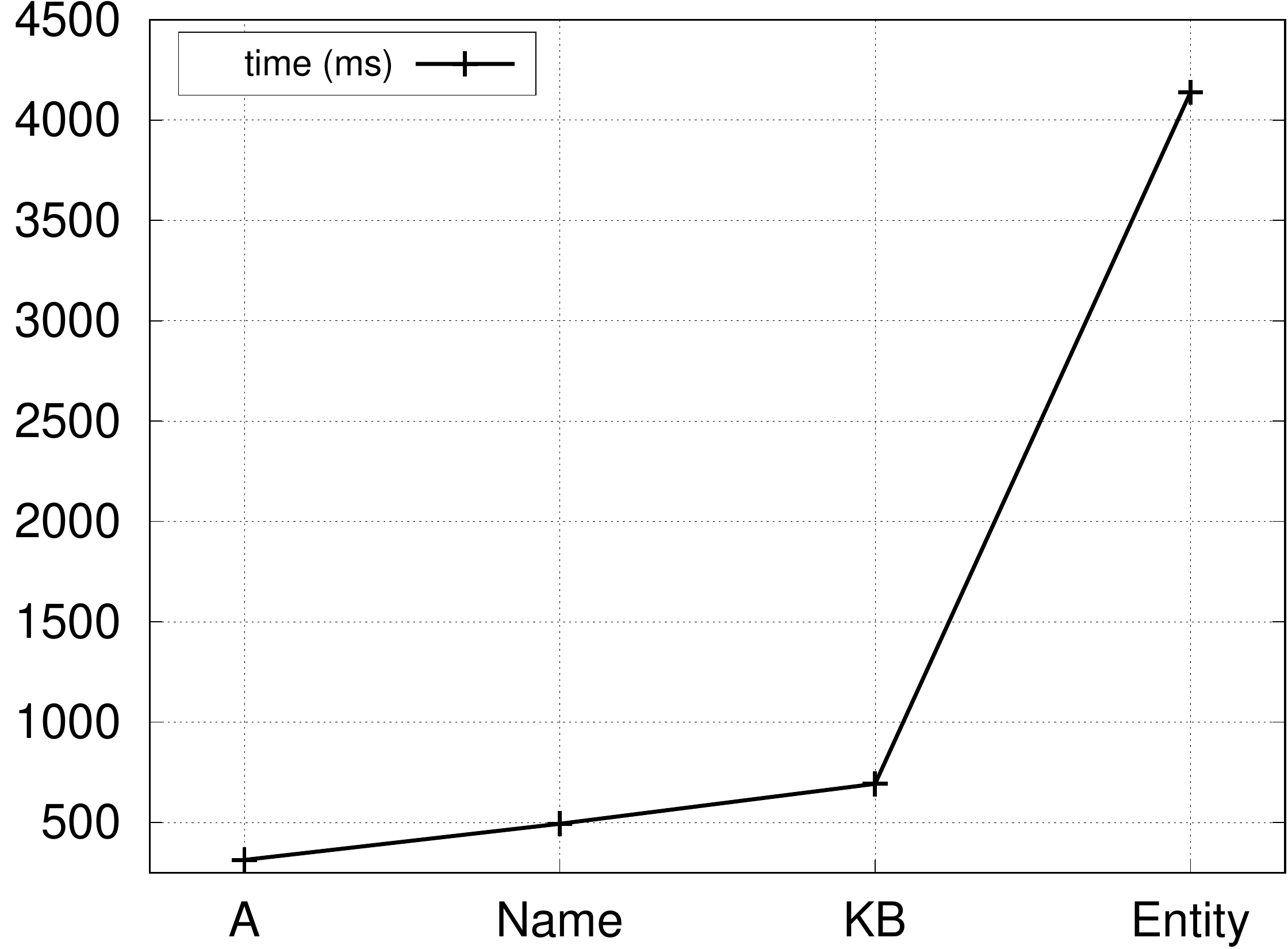}
     \caption{\NAME{}$_{aida}$: Timing experiments for CoNLL2003e-test in average milliseconds per document}
     \label{fig:timing-experiments}
\label{figure:timming}
 \end{figure}

However, the ablation study in Tab.~\ref{table:ablation} suggests that some KB-based features may be
subsumed by the Entity-based ones which generates the most significant boost. 
This is somehow expected as the entity specific information is extracted
from the same KB and strongly relies on the entity types. The Entity-based component is also the most expensive
concerning timing performance. Fig.~\ref{figure:timming} shows the time required by each setting,
establishing a trade-off between accuracy and runtime.
The Stanford agnostic system was faster than our implementation as
it took 158.55 ms per document on average.

\begin{table}[ht]
	\centering
    \begin{center}
		\scalebox{1}{
        \begin{tabular}{|l|c|}
            \hline
            \bf Feature Categories & $F_1$ \\
            \hline \hline
            A, Name, KB & 88.73 \\
            \hline
            A, Name, Entity & 89.32 \\
            \hline
            A, KB, Entity & 91.09 \\
            \hline
            All & 91.12 \\
            \hline
        \end{tabular}
		}
    \end{center}
    \caption{\label{table:feature-ablation-study} \NAME{}$_{gold}$: Ablation study by categories on CoNLL2003e-test}
 \label{table:ablation}
\end{table}

\begin{figure*}[!hbt]
     \centering

     \begin{subfigure}{0.48\textwidth}
         \includegraphics[width=1\linewidth,keepaspectratio]{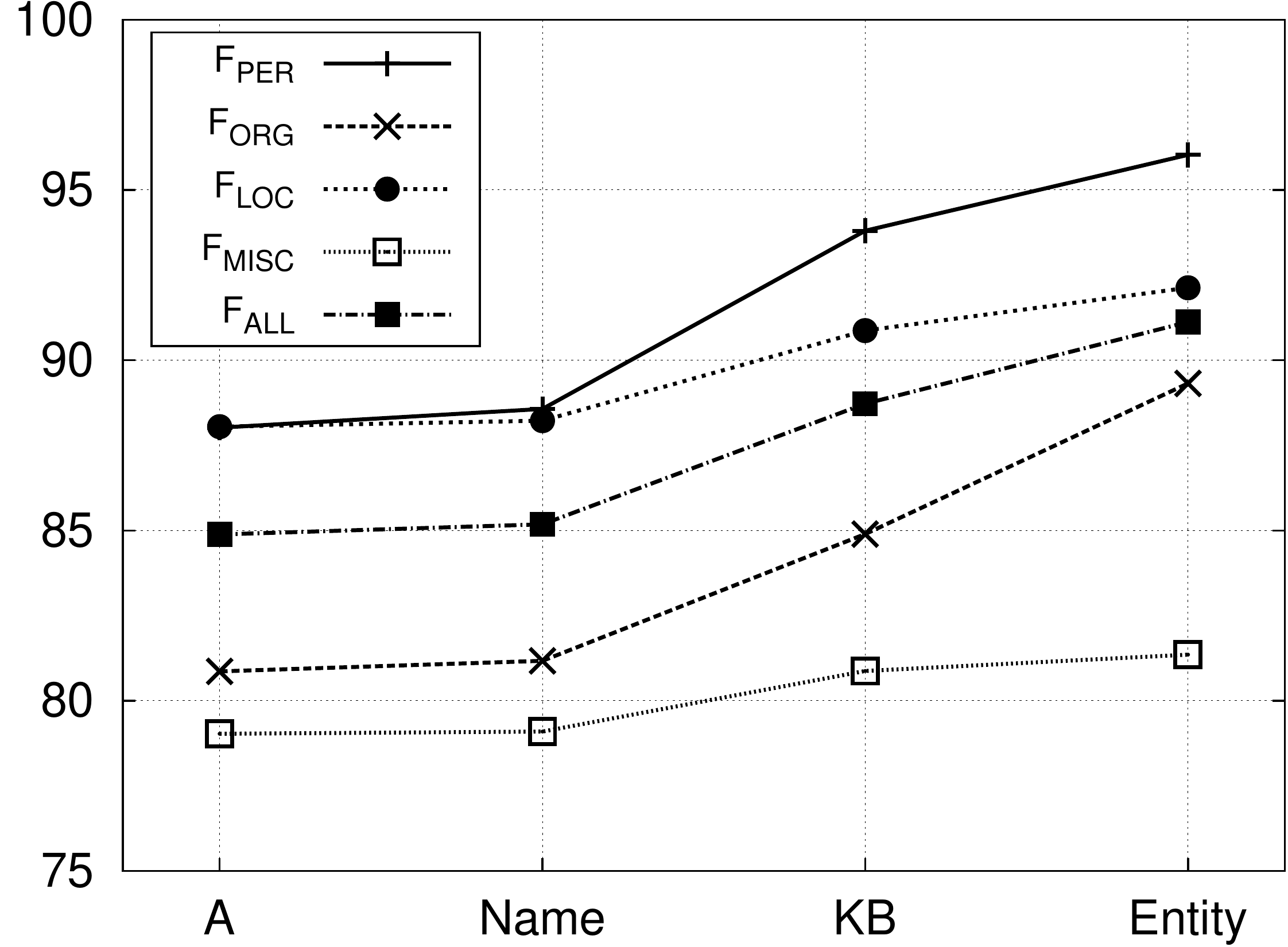}
         \caption{Mention-based $F_{1}$ score on CoNLL2003e-test}
         \label{fig:feature-categories-conll-test}
     \end{subfigure}
     \quad
     \begin{subfigure}{0.48\textwidth}
         \includegraphics[width=1\linewidth,keepaspectratio]{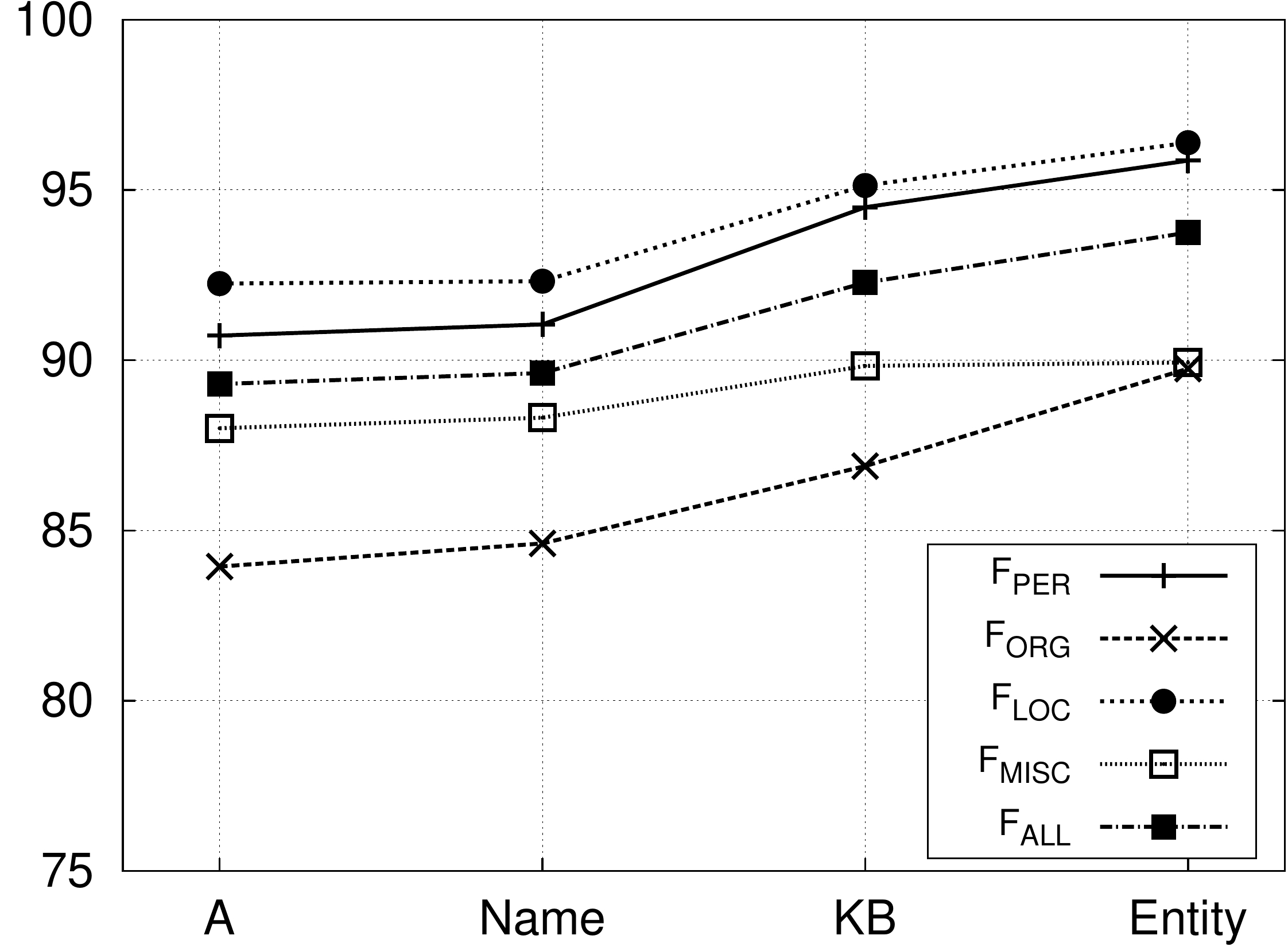}
         \caption{Mention-based $F_{1}$ score on CoNLL2003e-dev}
         \label{fig:feature-categories-conll-dev}
     \end{subfigure}

     \begin{subfigure}{0.48\textwidth}
         \includegraphics[width=1\linewidth,keepaspectratio]{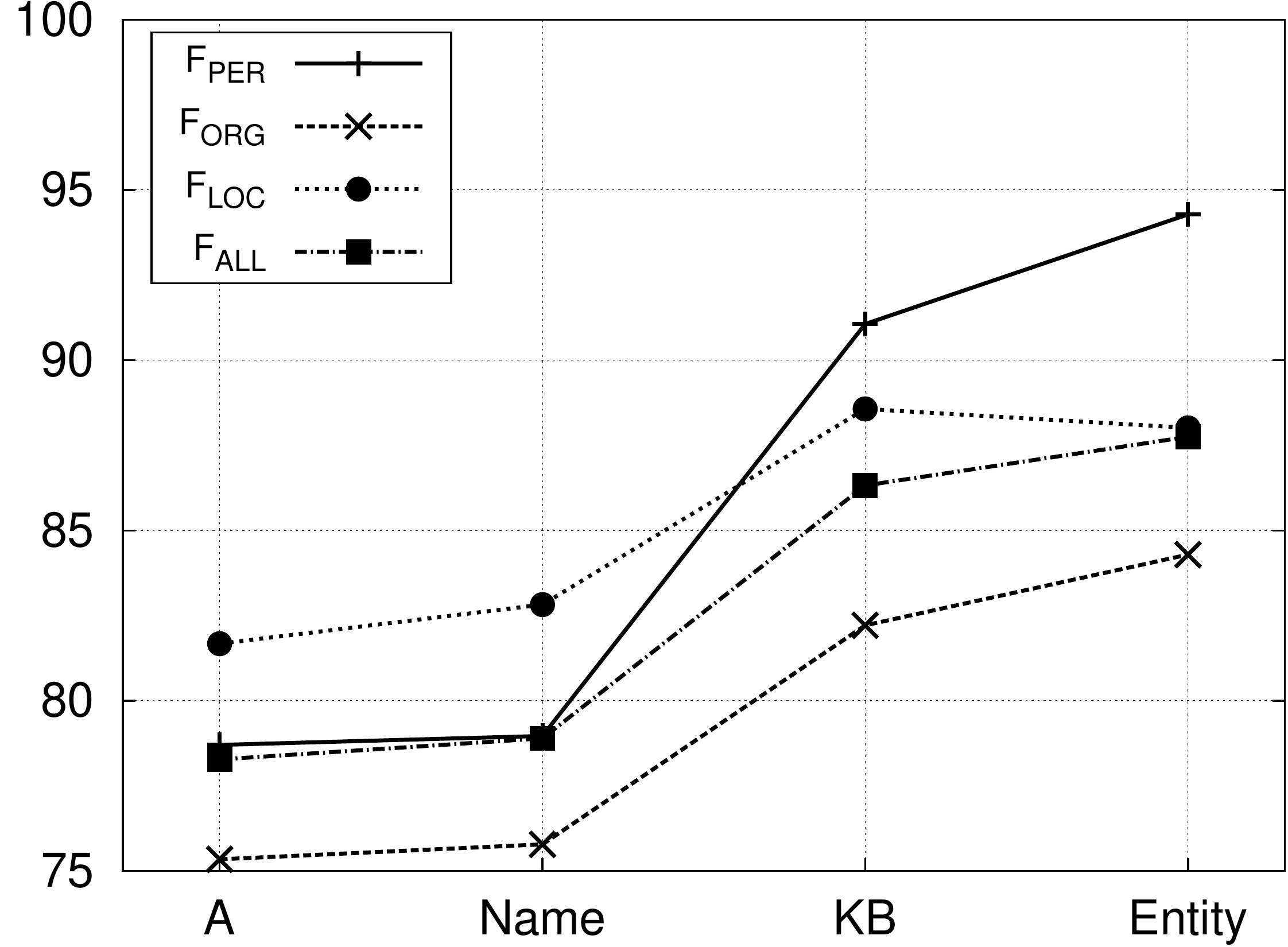}
         \caption{Mention-based $F_{1}$ score on MUC-7-test}
         \label{fig:feature-categories-muc-test}
     \end{subfigure}
     \quad
     \begin{subfigure}{0.48\textwidth}
         \includegraphics[width=1\linewidth,keepaspectratio]{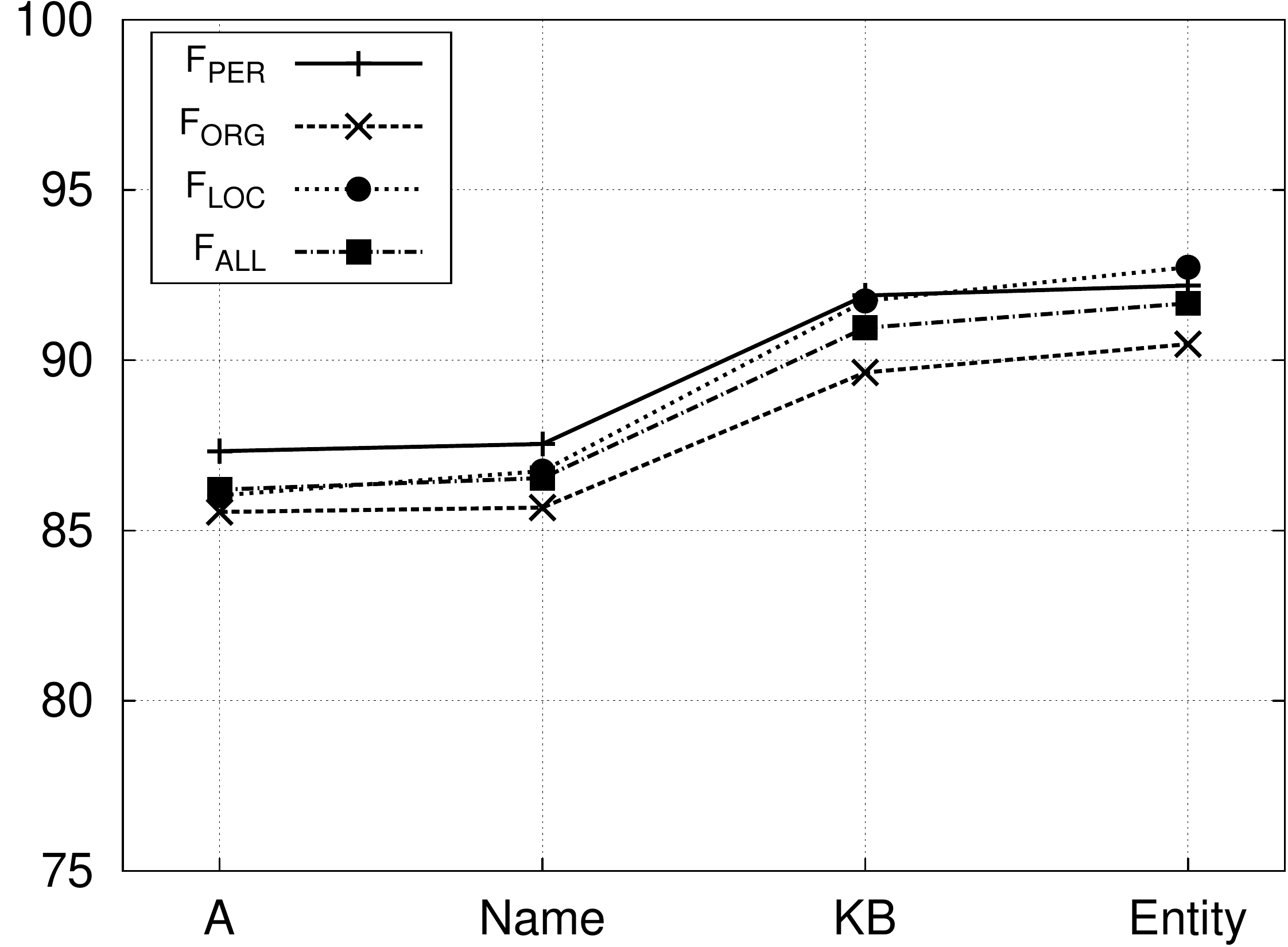}
         \caption{Mention-based $F_{1}$ score on MUC-7-dev}
         \label{fig:feature-categories-muc-dev}
     \end{subfigure}

     \caption{\NAME{}$_{gold}$: Incremental knowledge results}
 \end{figure*}

Fig.~\ref{fig:feature-categories-conll-test} and Fig.~\ref{fig:feature-categories-conll-dev} show the performance for each specific entity type for both CoNLL2003e-test and CoNLL2003e-dev. \NAME{} achieves human-level performance for labelling persons ($F_{1}$ 96.03 and $F_{1}$ 95.86) and locations ($F_{1}$ 92.13 and $F_{1}$ 96.39). The positive effect of external knowledge is quite significant for organizations ($F_{1}$ 80.86 to $F_{1}$ 89.32 on test; $F_{1}$ 83.94 to $F_{1}$ 89.75 on dev) while it is relatively moderate for miscellaneous. In the case of MUC-7 (Fig.~\ref{fig:feature-categories-muc-test} and Fig.~\ref{fig:feature-categories-muc-dev}), the effect is similar except for locations in MUC-7-test which tend to slightly drop when the entity-based category is used. The positive impact on persons for MUC-7-test is especially significative as it generates a change in ranking performance with respect to the other types. It jumps from the second and third position on MUC-7 test and MUC-7 dev with agnostic features to the very first position in MUC-7-test and the second on MUC-7-dev. 

Finally, Tab.~\ref{table:type-based-f1-muc} and Tab.~\ref{table:type-based-f1-conll}  display mention-based results on \\ CoNLL2003e and MUC-7 for all knowledge categories and entity types. They also display the span-based performance for each knowledge category. The numbers suggest that adding knowledge improves task performance.

\begin{table*}[!htb]
    \begin{center}
        \scalebox{1}{
            \begin{tabular}{|c|c|c|c|c|c|c|c|c|c|c|}
                \hline
                \bf \multirow{2}{*}{Cat.} & \multicolumn{5}{|c|}{\bf MUC-7 test} & \multicolumn{5}{|c|}{\bf MUC-7 dev} \\
                \cline{2-11}
                & $F_{PER}$ & $F_{ORG}$ & $F_{LOC}$ & $F_{ALL}$ & $F_{SPAN}$ & $F_{PER}$ & $F_{ORG}$ & $F_{LOC}$ &  $F_{ALL}$ & $F_{SPAN}$ \\
                \hline
                A & 78.70 & 75.34 & 81.67 & 78.27 & 84.75 & 87.33 & 85.54 & 86.03 & 86.20 & 88.96 \\
                Name & 78.96 & 75.78 & 82.82 & 78.89 & 84.48 & 87.54 & 85.67 & 86.74 & 86.54 & 88.86 \\
                KB & 91.06 & 82.21 & 88.57 & 86.32 & 86.75 & 91.90 & 89.64 & 91.74 & 90.95 & 90.80 \\
                Entity & 94.28 & 84.29 & 88.01 & 87.75 & 87.97 & 92.19 & 90.47 & 92.73 & 91.67 & 91.67 \\
                \hline
            \end{tabular}
        }
    \end{center}
    \caption{\NAME{}$_{gold}$: Mention $ F_{1} $ (knowledge category and type) and Span $ F_{1}$ (knowledge category) on MUC-7.}
    \label{table:type-based-f1-muc}
\end{table*}

\begin{table*}[!htb]
    \begin{center}
        \scalebox{1}{
            \begin{tabular}{|c|c|c|c|c|c|c|c|c|c|c|c|c|}
                \hline
                \bf \multirow{2}{*}{Cat.} & \multicolumn{6}{|c|}{\bf CoNLL2003e test} & \multicolumn{6}{|c|}{\bf CoNLL2003e dev} \\
                \cline{2-13}
                & $F_{PER}$ & $F_{ORG}$ & $F_{LOC}$ & $F_{MISC}$ & $F_{ALL}$ & $F_{SPAN}$ & $F_{PER}$ & $F_{ORG}$ & $F_{LOC}$ & $F_{MISC}$ & $F_{ALL}$ & $F_{SPAN}$ \\
                \hline
                A & 88.02 & 80.86 & 88.05 & 79.03 & 84.88 & 93.39 & 90.72 & 83.94 & 92.25 & 88.00 & 89.30 & 95.22 \\
                Name & 88.57 & 81.17 & 88.22 & 79.09 & 85.18 & 93.17 & 91.05 & 84.62 & 92.32 & 88.31 & 89.62 & 95.04 \\
                KB & 93.80 & 84.89 & 90.87 & 80.87 & 88.72 & 94.35 & 94.49 & 86.89 & 95.13 & 89.83 & 92.28 & 95.68 \\
                Entity & 96.03 & 89.32 & 92.13 & 81.35 & 91.12 & 94.82 & 95.86 & 89.75 & 96.39 & 89.93 & 93.75 & 96.38 \\
                \hline
            \end{tabular}
        }
    \end{center}
    \caption{\NAME{}$_{gold}$: Mention $ F_{1} $ (knowledge category and type) and Span $ F_{1}$ (knowledge category) on CoNLL2003e.}
    \label{table:type-based-f1-conll}
\end{table*}

\subsection{Comparative Performance}
\label{sec:comparative-performance}

{\bf English.} \NAME{} performs amongst state-of-the-art NER English systems on both datasets. Tab.~\ref{table:competitors} reports the results for mention-based performance on CoNLL2003e-test compared to the best-known systems. The results for \NAME{} correspond to a setting using all the knowledge categories when using the gold standard for the entity-based step (as in Radford at al.~\cite{RadfordCH15}) or using the AIDA system for the entity-based knowledge category.  

\begin{table}[!htb]
    \begin{center}
    	\scalebox{1}{
        \begin{tabular}{|c|c|}
            \hline
            \bf System & \bf $F_{1}$ \\
            \hline \hline
            Chiu and Nichols~\cite{Chiu:2016} & 91.62 \\ \hline
            Luo	et al.~\cite{LuoHLN15} & 91.20 \\ \hline
			Yang et al.~\cite{YangSC16} & 91.20 \\ \hline
            \bf \NAME{}$_{gold}$ & \bf 91.12 \\ \hline
			Lample et al.~\cite{LampleBSKD16} & 90.94 \\ \hline
            Passos et al.~\cite{PassosKM14} & 90.90 \\ \hline
            Lin and Wu~\cite{LinW09} & 90.90 \\ \hline
            Ratinov and Roth~\cite{RatinovR09} & 90.80 \\ \hline
            \bf \NAME{}$_{aida}$ & \bf 90.16 \\ \hline
            Radford et al.~\cite{RadfordCH15} & 89.35 \\ \hline
            Finkel et al.~\cite{FinkelGM05} & 86.86 \\ \hline
        \end{tabular}
    }
    \end{center}
    \caption{\label{table:competitors} English: Mention-based performance on CoNLL2003e-test as reported in literature.}
\end{table}

Tab.~\ref{table:stanford-muc} displays detailed results for one of the latest versions of Finkel et al.\cite{FinkelGM05} (Stanford NER 3.6.0), probably the most widely used NER system to date, which \NAME{} outperforms.

\begin{table}[!htb]
    \begin{center}
        \scalebox{1}{
            \begin{tabular}{|c|l|c|c|}
                \hline
                System & Type & \specialcell{CoNLL2003e\\test} & \specialcell{CoNLL2003e\\dev} \\\hline
                \multirow{5}{1.5cm}{Stanford \\[.2\baselineskip] CoreNLP (3.6.0)} & LOC     & 89.04 & 94.38 \\
                                          & MISC    & 81.51 & 87.44 \\
                                          & ORG     & 85.62 & 88.33 \\
                                          & PER     & 92.35 & 94.28 \\
                                          & All     & 88.05 & 91.95 \\\hline\hline	
                \multirow{5}{*}{\NAME{}$_{aida}$} & LOC     & 91.28 & 95.33 \\
                                          & MISC    & 81.82 & 88.59 \\
                                          & ORG     & 87.43 & 88.82 \\
                                          & PER     & 95.99 & 95.75 \\
                                          & All     & 90.16 & 93.11 \\\hline\hline					  
                \small \NAME{}$_{gold}$ & All & 91.12 & 93.75 \\\hline
            \end{tabular}
        }
	    \end{center}
        \caption{\label{table:stanford-muc} English: F$_{1}$ Performance for the English language on CoNLL2003e and MUC-7 datasets.}
\end{table}

{\bf German.} To the best of our knowledge, \NAME{} is one of the best performing systems to date for the German language on CoNLL2003g. Tab.~\ref{table:competitors-german} presents the results for \NAME{} compared with state-of-the-art systems. Tab.~\ref{table:german-types} presents detailed results for each named entity type. The biggest boost in Germany is generated by the entity-based features, which generate an increment of more than 7 points in recall with respect to the previous knowledge category (i.e., kb-based).

\begin{table}[!htb]
    \begin{center}
    	\scalebox{1}{
        \begin{tabular}{|c|c|}
            \hline
            \bf System & \bf $F_{1}$ \\
            \hline \hline
 			Lample et al.~\cite{LampleBSKD16} & 78.76 \\ \hline
            \bf \NAME{}$_{aida}$ & \bf 77.20 \\ \hline
            Gillick et al.~\cite{GillickBVS16} & 76.22 \\ \hline
            Qi et al.~\cite{QiCKKW09} & 75.72 \\ \hline
        \end{tabular}
    }
    \end{center}
    \caption{\label{table:competitors-german} German: Mention-based performance on CoNLL2003g (German) as reported in literature.}
\end{table}

\begin{table}[!htb]
    \begin{center}
        \scalebox{1}{
            \begin{tabular}{|l|c|c|}
                \hline
                Type & \specialcell{CoNLL2003g\\test} & \specialcell{CoNLL2003g\\dev}   \\\hline
                LOC     & 77.24 & 78.03  \\
                MISC    & 68.79 & 72.59 \\
                ORG     & 65.58 & 75.65 \\
                PER     & 88.59 & 90.21 \\
                All     & 77.20 & 79.93 \\\hline\hline					  
            \end{tabular}
        }
	    \end{center}
        \caption{\label{table:german-types} German: \NAME{}$_{aida}$ F$_{1}$ Performance for the German language on CoNLL2003g dataset.}
\end{table}

{\bf Spanish.} Tab.~\ref{table:competitors-spanish} presents the results for \NAME{} compared with state-of-the-art systems for Spanish. Tab.~\ref{table:spanish-types} presents detailed results for each named entity type. 

\begin{table}[!htb]
    \begin{center}
    	\scalebox{1}{
        \begin{tabular}{|c|c|}
            \hline
            \bf System & \bf $F_{1}$ \\
            \hline \hline
            Yang et al.~\cite{YangSC16} & 85.77 \\ \hline
            Lample et al.~\cite{LampleBSKD16} & 85.75 \\ \hline
            \bf \NAME{}$_{aida}$ & \bf 83.79 \\ \hline
            Gillick et al.~\cite{GillickBVS16} & 82.95 \\ \hline
			dos Santos and Guimar{\~{a}}es~\cite{SantosG15} & 82.21\\ \hline
        \end{tabular}
    }
    \end{center}
    \caption{\label{table:competitors-spanish} Spanish: Mention-based performance on CoNLL2002 as reported in literature.}
\end{table}

\begin{table}[!htb]
    \begin{center}
        \scalebox{1}{
            \begin{tabular}{|l|c|c|}
                \hline
                Type & \specialcell{CoNLL2002\\test} & \specialcell{CoNLL2002\\dev}   \\\hline
                LOC     & 83.92 & 81.18  \\
                MISC    & 59.19 & 55.15 \\
                ORG     & 83.03 & 80.79 \\
                PER     & 94.34 & 93.48 \\
                All     & 83.79 & 82.14 \\\hline				  
            \end{tabular}
        }
	    \end{center}
        \caption{\label{table:spanish-types} Spanish: \NAME{}$_{aida}$ F$_{1}$ Performance for the Spanish language on CoNLL2002 dataset.}
\end{table}

\section{Related Work}

NER is a widely studied problem in the natural language understanding community. Well developed work has established a clear direction towards the use of CRFs~\cite{Lafferty01} with systems achieving high relative performance~\cite{FinkelGM05,KazamaT07a,RatinovR09,PassosKM14,RadfordCH15,LuoHLN15}. A new line, focused on neural networks methods~\cite{SantosG15,Chiu:2016,LampleBSKD16,YangSC16,YangSC16,GillickBVS16}. Chiu and Nichols~\cite{Chiu:2016}, for instance, the best NER system for English to date implemented a hybrid bidirectional LSTM-CNN whose inputs are tokens, word and character embeddings, and a set of  gazetteers with type encodings.

Among the CRF methods, early work has focused on purely agnostic systems~\cite{Klein:2003,FinkelGM05}. Klein et al.~\cite{Klein:2003} presents a system addressing the importance of substring features, an idea that we also capture in our agnostic model via prefixes and suffixes. Finkel et al.~\cite{FinkelGM05} is one of the most popular agnostic systems. Following this work, our agnostic category implements most of the features described in the paper plus prefixes and suffixes used in Zhang and Johnson~\cite{Zhang03}. We do not make use of feature type statistics from the dataset which may explain a small drop in performance in our agnostic setting compared to Finkel et al.~\cite{FinkelGM05} in our English experiments. In contrast to agnostic approaches, our system strongly relies on background knowledge to improve performance.

Previous work has already regarded NER as a knowledge intensive task~\cite{FlorianIJ003,Zhang03,KazamaT07a,RatinovR09,LinW09,PassosKM14,RadfordCH15,LuoHLN15}. Most of these works incorporate background knowledge in the form of entity-type gazetteers~\cite{FlorianIJ003,Zhang03,KazamaT07a,RatinovR09,PassosKM14}. In fact, dictionaries were already provided for the early CoNLL2003 shared-task encouraging the use of external knowledge. Ratinov and Roth~\cite{RatinovR09} used 30 gazetteers mostly extracted from Wikipedia, thereby generating big boosts in performance. These gazetteers have been successfully reused by other systems~\cite{PassosKM14,RadfordCH15,LuoHLN15}. In particular, Luo et al.~\cite{LuoHLN15} used a total of 655 gazetteers including those from Ratinov and Roth~\cite{RatinovR09}. We also incorporate gazetteers in our knowledge-based features. Finally, Kazama and Torisawa~\cite{KazamaT07a} was one of the first works to extract type information from Wikipedia. Their approach extracts category labels from the first sentence of the Wikipedia entity pages. In our method we also explicitly incorporate type information in the KB-based and the entity-based categories but in a cleaner way as they are derived from a high precision knowledge base like YAGO.

Compared to previous approaches using external knowledge, our method is more modular in the way the knowledge is incorporated. Our framework allows us to classify and easily derive more features. We have both more light-weight and knowledge-intensive features from different sources: entity names, knowledge bases and NED. Our distinctive features include frequent mention tokens, frequent mention shapes, frequent POS mention patterns, Wikipedia token probability and the class type probability, among others (Sec.~\ref{sec:knowledge-aug-ner}).  

The association between NER and NED has been successfully exploited by recent work~\cite{DurrettKlein2014,RadfordCH15,LuoHLN15} as a means to boost NER performance. Radford et al.~\cite{RadfordCH15} uses a two-step approach. They showed that local features derived from an initial NED run improve the performance on a second NER step. Specifically, alternative entity names and types tend to be important. We follow a similar two-step approach but, in contrast, we also run NED using a real world setting. Luo et al.~\cite{LuoHLN15} present a joint model for named entity recognition and disambiguation, as a CRF with a topology for joint optimization.  NER and NED tasks are inherently associated so performing these tasks jointly poses natural advantages. However, NER has multiple applications apart from NED, which tends to be a computationally expensive task. Our modular approach, on top of a simpler and more tractable model, avoids expensive joint optimisation and permits to easily decouple the NED module for settings with small computational requirements. It also benefits from the mutual dependency between NER and NED when heavy computation is not an issue.

Regarding multilinguality, recent work has focused on methods to handle NER across a wide set of languages~\cite{YangSC16,LampleBSKD16,GillickBVS16}. Yang et al.~\cite{YangSC16}, one of the best systems across languages, implements a hierarchical recurrent neural network for joint POS tagging, chunking and NER, implemented on top of a CRF layer to do the labelling.

\section{Conclusion}

We presented \NAME, a multilingual system that explicitly encodes different degrees
of external knowledge for NER. \NAME's framework defines four knowledge categories,
each containing deeper external knowledge. 
Our experimental study shows that \NAME\ performs among state-of-the-art NER systems
across languages.  It also shows that increasing the degree of external knowledge 
encoded in the system significantly boosts NER performance.

\bibliographystyle{ACM-Reference-Format}
\bibliography{sigproc} 

\end{document}